\documentclass[letterpaper, 10 pt, journal, twoside]{IEEEtran}
\usepackage{cite}
\usepackage{amsmath,amssymb,amsfonts}
\usepackage{algorithmic}
\usepackage{graphicx}
\usepackage{textcomp}
\usepackage{tabularx}  % for 'tabularx' environment and 'X' column type
\usepackage{ragged2e}  % for '\RaggedRight' macro (allows hyphenation)
\usepackage{makecell}  % for '\makecell' macro
\usepackage{array} % Add this line to your preamble
\usepackage{mdframed}
\usepackage{comment}
\usepackage{url}
\usepackage[T1]{fontenc}
\usepackage{csquotes}
\usepackage{float} % For precise float placement with [H] specifier
\usepackage{tabularx}  % For 'tabularx' environment
\usepackage{booktabs}  % For \toprule, \midrule, \bottomrule
\usepackage{multirow}  % For \multirow command
\usepackage{hyperref}
\usepackage{caption}
\usepackage[table]{xcolor} 

\newcolumntype{L}{>{\RaggedRight\arraybackslash}X} % modified 'X' column type

\def\BibTeX{{\rm B\kern-.05em{\sc i\kern-.025em b}\kern-.08em
    T\kern-.1667em\lower.7ex\hbox{E}\kern-.125emX}}

\IEEEoverridecommandlockouts
\title{A Taxonomy of Self-Handover}
\author{
Naoki Wake$^{1}$,
Atsushi Kanehira$^{1}$,
Kazuhiro Sasabuchi$^{1}$,
Jun Takamatsu$^{1}$,
and Katsushi Ikeuchi$^{1}$
\thanks{$^{1}$Applied Robotics Research, Microsoft, Redmond, WA 98052, USA
        {\tt\footnotesize naoki.wake@microsoft.com}}%
\thanks{Digital Object Identifier (DOI): see top of this page.}
}
\begin{document}
\maketitle
\begin{abstract}
Self-handover, transferring an object between one's own hands, is a common but understudied bimanual action. While it facilitates seamless transitions in complex tasks, the strategies underlying its execution remain largely unexplored. Here, we introduce the first systematic taxonomy of self-handover, derived from manual annotation of over 12 hours of cooking activity performed by 21 participants. Our analysis reveals that self-handover is not merely a passive transition, but a highly coordinated action involving anticipatory adjustments by both hands. As a step toward automated analysis of human manipulation, we further demonstrate the feasibility of classifying self-handover types using a state-of-the-art vision-language model. These findings offer fresh insights into bimanual coordination, underscoring the role of self-handover in enabling smooth task transitions---an ability essential for adaptive dual-arm robotics.
\end{abstract}


\section{Introduction}
\IEEEPARstart{H}{umans} skillfully perform coordinated bimanual actions in everyday life. Among them, \textit{self-handover}---transferring an object between one’s own hands without intermediate placement---is remarkably common, yet largely overlooked~\cite{krebs2022bimanual,nakamura2019effect}. We define self-handover as the transition from holding an object with one hand to either passing it to the other hand or engaging both hands in manipulation. This seemingly simple action restructures the spatial configuration of hands, objects, and their surroundings---elements that fundamentally shape how manipulation unfolds~\cite{ikeuchi2024applying}. In doing so, self-handover acts as a hinge between subtasks, enabling seamless and adaptive task flow.

Bimanual manipulation has long been studied in robotics and neuroscience~\cite{smith2012dual,swinnen2004two}, with recent advances in hardware and learning-based planning renewing interest in developing dexterous, coordinated skills~\cite{krebs2022bimanual,rakita2019shared,vezzani2017novel,pavlichenko2019autonomous,colombo2024manipulation,wan2016achieving,mousavi2013new}. Within this context, human demonstrations have served as a rich source of insight for robotic systems~\cite{wake2023gpt,ikeuchi2024semantic,krebs2022bimanual,zhou2024learning}. Notably, Krebs and Asfour proposed a comprehensive taxonomy of bimanual manipulation, describing actions in terms of inter-hand coordination, spatial constraints, and role division~\cite{krebs2022bimanual}. While self-handover may be subsumed under this framework, its specific patterns and functions remain unexamined.

\begin{figure}[t]
  \centering
  \includegraphics[width=0.48\textwidth]{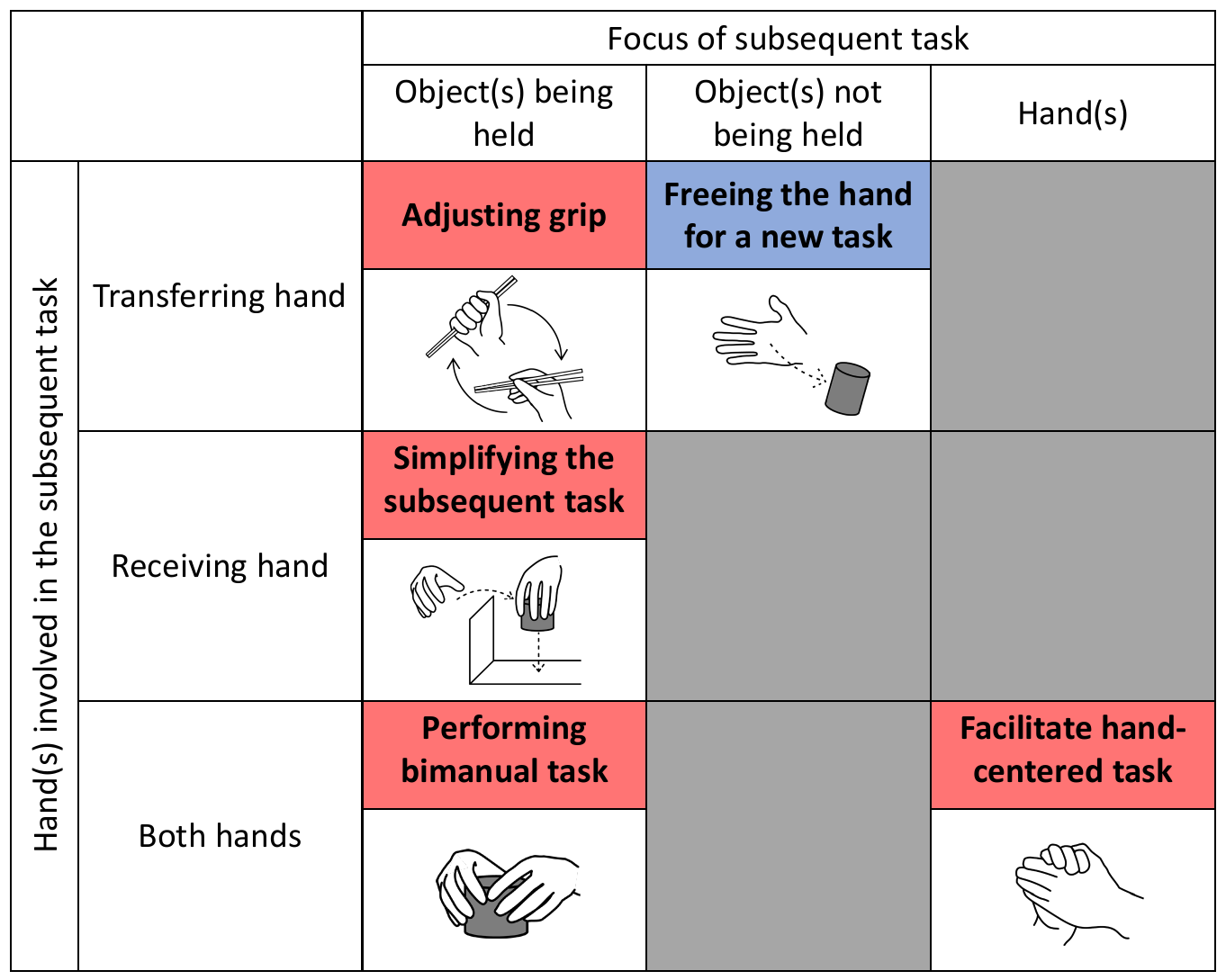}
  \caption{A functional taxonomy of self-handover behaviors. Self-handover---the transfer of an object between one’s own hands---takes on different forms depending on the goal of the handover and the roles of each hand. We identify six core patterns, structured by whether the handover targets an object or a hand, and whether one or both hands remain active afterward. Red cells indicate active self-handovers, where the object plays a central role in the next action. Blue marks passive transitions, where the object is temporarily released to free a hand. Full definitions of each category are provided in Section~\ref{taxonomy}.}
  \label{fig:taxonomy}
\end{figure}

Here, we systematically investigate self-handover in naturalistic human behavior. Focusing on cooking---a domain rich in diverse manipulations---we analyze over 12 hours of egocentric video from 21 participants, extracting and annotating self-handover episodes. From these, we construct a taxonomy of self-handover patterns and examine their roles in task progression (Fig.~\ref{fig:taxonomy}).

To explore automated recognition, we further present a classification approach using a Vision-Language Model (VLM), leveraging recent progress in multimodal learning. While challenges remain—particularly in reducing hallucinations—our findings indicate that VLMs can capture meaningful aspects of human intent related to self-handover.

Our contributions are threefold:
\begin{itemize}
    \item A taxonomy for categorizing self-handover behaviors in naturalistic tasks;
    \item A functional analysis of human self-handover strategies and their impact on task efficiency;
    \item A demonstration of self-handover recognition using a VLM.
\end{itemize}

Through this study, we aim to deepen the understanding of how self-handover supports fluid and efficient task execution, and to advance models of bimanual coordination in both humans and robots.

\section{Related work}
\subsection{Bimanual Manipulation in Humans}~\label{human_bimanual}
The understanding of bimanual manipulation in humans has a long history and has been extensively studied in neuroscience and related fields. Much of the research has focused on rhythmic coordination tasks, where both arms perform similar or synchronized movements. Such synchronized movements are typically studied from an information-processing standpoint, focusing on how limited neural resources control both limbs~\cite{swinnen2004two}. In contrast, goal-directed bimanual manipulations also exist, where the two arms exert distinct but coordinated motor outputs to achieve a common objective (e.g., buttoning a shirt). Self-handover is one such goal-directed behavior, yet its underlying neural mechanisms remain poorly understood.~\cite{liao2018neural,obhi2004bimanual,guiard1987asymmetric}. Although one hand’s movement can interfere with the other~\cite{osumi2017restoring}, how the human brain coordinates bimanual manipulation remains an open question.

From a behavioral science perspective, bimanual manipulation has been studied primarily in the context of reaching and pick-and-place tasks that involve self-handover. Gonzalez et al.~\cite{gonzalez2015hand} reported a general tendency for people to use the right hand for objects located on the right side, and vice versa, during reaching tasks. Nakamura et al.~\cite{nakamura2019effect} examined object transportation tasks and demonstrated that self-handover occurs more frequently when transporting objects between the left and right sides than during unimanual transport or simultaneous bimanual use. These findings suggest that hand choice---and the decision to perform a self-handover---depends on the spatial relationship between body and environment~\cite{rosenbaum2010psychologically}. Such strategies indicate that humans efficiently coordinate both arms to optimize task performance. Our work extends these findings by investigating self-handover behaviors beyond reaching and pick-and-place tasks, covering a broader range of actions that occur during cooking. We further aim to develop a taxonomy that systematically categorizes these behaviors.

\subsection{Bimanual Manipulation in Robots}~\label{robot_bimanual}
Robotic bimanual manipulation has been studied for decades, with early robotic systems including two-arm configurations, often controlled via teleoperation~\cite{ambrose2000robonaut}. While single-arm robots dominated industrial applications for a long time, the advancement of humanoid hardware and learning techniques has reignited interest in dual-arm robotic systems since the 2000s~\cite{smith2012dual}. The primary advantage of bimanual robots lies in their high task-space redundancy~\cite{almeida2018cooperative,ogren2012multi}. However, the use of two arms also introduces significant challenges, including redundant degrees of freedom, inter-arm interference~\cite{wang2014dual}, complex kinematic coordination~\cite{wang2015cooperative}, force control~\cite{stepputtis2022system}, and high-level task planning~\cite{gao2024dag,chu2024large}. Bimanual coordination remains an active trend in robotic manipulation research~\cite{billard2019trends}.

In recent years, imitation learning~\cite{zhao2023learning,motoda2025learning,lu2024anybimanual,lee2015learning,stepputtis2022system,grannen2023stabilize,kim2024goal,black2024pi_0} and reinforcement learning~\cite{chen2023bi,chitnis2020efficient,kroemer2015towards} have become increasingly prominent for training bimanual skills. While several learning-based models can execute bimanual tasks that include self-handover behaviors, they typically rely on explicit language inputs such as ``Hand over the \underline{\phantom{xxx}} item.~\cite{lu2024anybimanual}'' In contrast, humans perform self-handover adaptively in response to environmental and task demands, revealing a gap between current robotic approaches and human behavior. Although several control-based approaches have explicitly addressed self-handover---primarily focusing on regrasping or specific hand-switching scenarios~\cite{vezzani2017novel,pavlichenko2019autonomous,colombo2024manipulation,mousavi2013new,wan2016achieving,saut2010planning,balaguer2012bimanual}---a comprehensive method for achieving generalizable self-handover remains underexplored. By observing and categorizing human self-handover strategies, this study seeks to inform the design of robotic systems that perform such behaviors more efficiently and adaptively.

\subsection{Taxonomy of bimanual manipulation}
In the robotics community, taxonomies are a commonly used approach for analyzing and synthesizing complex actions. Well-designed taxonomies of domain-specific actions are widely applied for various purposes, such as analyzing human behavior~\cite{feix2014analysis,meixner2023evaluation}, learning robotic skills from human demonstrations~\cite{zollner2004programming,saito2022task}, and designing appropriate skill granularity for robotic systems~\cite{lee2024human,jaquier2022bringing,korsah2013comprehensive}. Targeted domains include grasping~\cite{arapi2021understanding,kamakura1980patterns,cutkosky1989grasp,feix2015grasp,liu2014taxonomy}, manipulation~\cite{paulius2020motion,krebs2022bimanual,borras2015whole,bullock2012hand,blanco2024t,park2016dual,ikeuchi2024semantic,boehm2021online}, and human-robot interaction~\cite{rodrigues2023multidimensional,yan2024hug}.

Taxonomies aim to provide a comprehensive description of actions by analyzing patterns derived either from theoretical frameworks or empirical observations. For example, in the domain of grasping, Yoshikawa’s closure theory offers a theoretical classification of force control in multifingered robot hands~\cite{yoshikawa2010multifingered}. Feix et al. proposed a taxonomy of human grasping patterns based on factors such as the functional grouping of fingers—referred to as virtual fingers~\cite{cutkosky1989grasp}—the involvement of the palm, thumb opposition types, and the force dynamics between the hand and the object~\cite{feix2015grasp}.

In manipulation, we have previously defined manipulation tasks as operations that alter the constraints imposed on objects and applied Kuhn-Tucker theory~\cite{kuhn1957linear} to derive a minimal and sufficient classification space based on linear inequality constraints~\cite{ikeuchi2024semantic}. While some work addresses bimanual manipulation~\cite{ureche2018constraints,volkmar2019improving,arapi2021understanding,bullock2012hand}, few taxonomies specifically focus on it. Krebs et al.~\cite{krebs2022bimanual} proposed a comprehensive taxonomy of bimanual manipulation aimed at supporting both the analysis of human motion and the execution of bimanual tasks by robots. Their taxonomy classifies bimanual manipulation according to the presence of inter-arm coordination, spatiotemporal dependencies between hands, and motion symmetry. They also demonstrated that rule-based automatic classification on everyday human activity datasets could reasonably identify categories within their taxonomy.

Despite the value of these taxonomies, a dedicated classification for self-handover remains underexplored. For instance, in Krebs et al.'s taxonomy, self-handover is treated as a single category, but the diverse patterns it includes have not been thoroughly examined. Mousavi et al.~\cite{mousavi2013new} proposed a taxonomy focused on regrasping, a subset of self-handover, but their framework does not comprehensively address the full range of self-handover behaviors. We aim to develop a dedicated taxonomy for self-handover, designed to support dual-arm robotic manipulation, with clearly distinguishable categories. As in Section~\ref{human_bimanual}, behavioral studies of humans suggest that arm coordination during self-handover serves to optimize the efficiency of subsequent tasks, indicating that facilitating the next action is a primary goal of self-handover. Based on this insight, we propose a taxonomy that classifies self-handover behaviors according to the semantic goal of the subsequent action, with a focus on the functional roles of the involved hands and objects. The presented taxonomy clarifies how humans employ self-handover to restructure tasks, potentially informing future research toward adaptive robotic bimanual manipulation.

\section{A Taxonomy of Self-Handover}\label{taxonomy}
This section details the proposed taxonomy of self-handover (Fig.~\ref{fig:taxonomy}). Based on insights from the existing literature discussed in Section\ref{human_bimanual}, self-handovers are considered to facilitate the execution of subsequent tasks. Accordingly, we propose to classify self-handovers from the perspective of which hand plays the primary role and what object is involved in the subsequent tasks. 
This classification helps determine whether a self-handover should be performed in a given scene, and if so, which type is most appropriate for the upcoming task.

Specifically, we classify self-handovers along two axes. The first axis concerns which hand serves as the main actor in executing the subsequent task: (1) the hand transferring the object, (2) the hand receiving the object, or (3) both hands. The second axis concerns the focus of the task: (1) the object being handed over, (2) an object other than the one being handed over, or (3) the hand(s) itself.

Our analysis of human cooking behavior, presented in Section~\ref{cooking}, revealed that actual instances of self-handover fall into five distinct categories within this $3 \times 3$ framework. The following subsections provide detailed explanations of each category.

\subsection{Adjusting Grip}
\begin{figure}[t]
  \centering
  \includegraphics[width=0.48\textwidth]{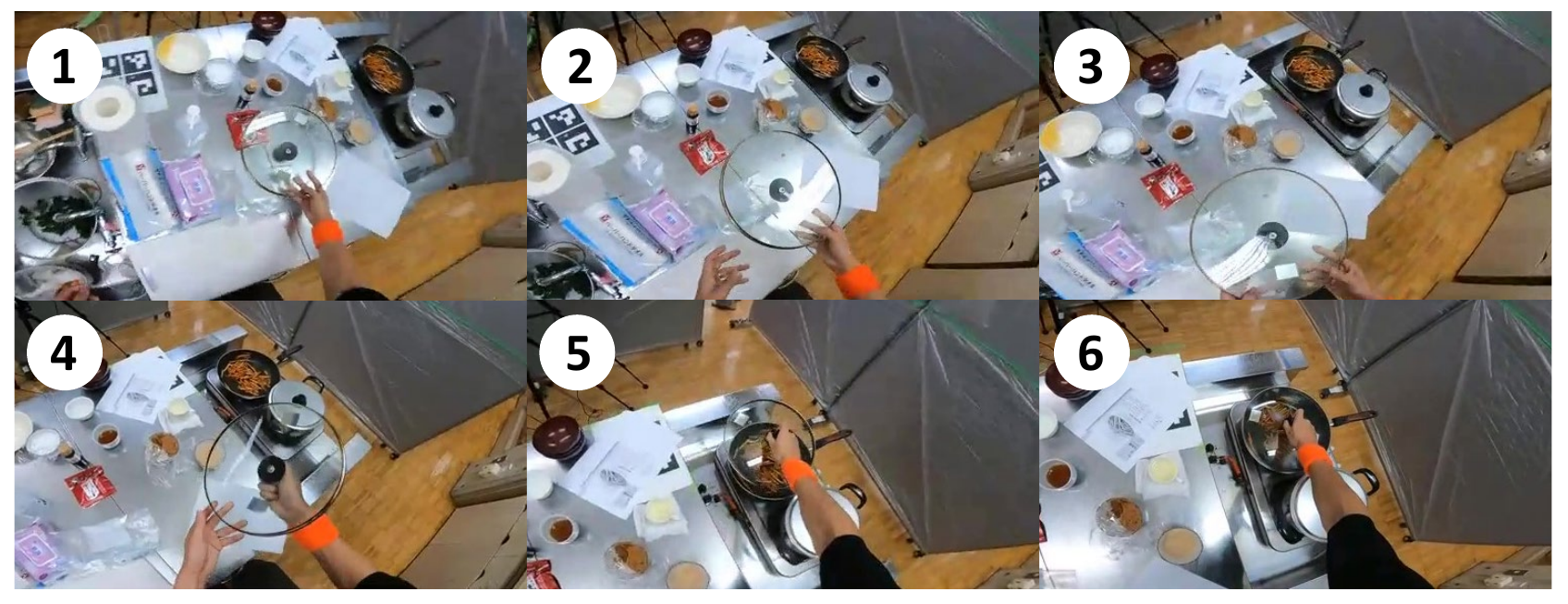}
  \caption{An example of self-handover for adjusting grip. The numbers indicate the chronological order of frames.}
  \label{fig:adjustgrip}
\end{figure}

This type of self-handover, also known as \textit{functional regrasping}~\cite{pavlichenko2019autonomous}, involves transferring an object to reconfigure the grasp for improved manipulation. For example, a person may reorient a bottle in-hand before opening it. In robotics, this behavior has been widely studied as a means of enabling dexterous manipulation~\cite{pavlichenko2019autonomous,colombo2024manipulation, wan2016achieving, mousavi2013new}. The goal here is not merely to change the holding hand, but to optimize the grasp configuration to meet a task-specific constraint. For example, one might pick up a lid from a table, pass it to the other hand, and then pass it back again to achieve a grip that is better suited for placing it over a frying pan (Fig.~\ref{fig:adjustgrip}).

Notably, functional regrasping can be achieved either through a single self-handover or through a back-and-forth exchange. While the former case, where the object is passed only once, may formally fall into the category of “Simplifying the Subsequent Task” according to our classification rules, the goals of self-handover in that category are diverse and not necessarily limited to regrasping. In contrast, we observed that the latter case, in which the object is passed back and forth, consistently exhibits a clear intent to adjust the grasp configuration. Accordingly, we classify the back-and-forth self-handover used to adjust grasp configuration as a distinct category, referred to as ``Adjusting Grip.''

\subsection{Simplifying the Subsequent Task}
In this category, self-handover is performed with the aim of improving the efficiency of subsequent tasks involving the object being passed. Efficiency here can refer to various factors, such as ease of posture when performing the action with the receiving hand, reduced physical movement (and thus energy consumption), minimized interference with the surrounding environment, or the use of the dominant hand. While using the dominant hand can be a motivation in itself, in many cases, the spatial relationship between the body, the environment, and the object makes self-handover a strategically beneficial choice.

An instance of this type of self-handover is cross-body transport, where an object is passed from one hand to the other across the body, as investigated by Nakamura et al.~\cite{nakamura2019effect}. Although such exchanges may appear to involve long-distance movement, tasks in this category are often much more localized. For example, self-handover is frequently observed even when the subsequent task lies within the reachable area of both hands---such as passing a ladle to the dominant hand before stirring, or transferring an object to place it onto a tray.

Interestingly, in cases where long-distance transport is involved, our observations suggest that humans often perform self-handover well in advance, possibly considering the spatial and temporal structure of the task. For instance, when carrying a rice bowl from the stove to the sink, a self-handover occurred before the actual transport begins.

\subsection{Performing a Bimanual Task}
This type of self-handover transitions into a dexterous bimanual task after the receiving hand grasps the object.
The receiving hand engages with the object with the aim of performing a task that is either difficult or less efficient to complete with one hand alone.
Bimanual tasks observed in this context are diverse. Examples include rinsing vegetables with water, shaking wet vegetables to remove excess water, folding or tying deformable objects such as plastic bags, and holding an object in a stable posture---such as placing a pot or cutting board horizontally on a table, or putting on a pair of glasses. Some tasks involve manipulating a single object in a way that conceptually divides it into two, or vice versa, from the perspective of manipulation.

Among these, we frequently observed a subset of bimanual tasks that involve the separation of one object into two, or the combination of two objects into one. Specific examples include peeling apart overlapping slices of meat using both hands; gripping a bell pepper sliced in half lengthwise to remove the seeds; and separating a stack of printed recipes. Notably, the objects involved in separation or combination are not always of the same type. For instance, we observed rotating the cap of a small jar to detach it from the container, peeling or applying plastic wrap to a dish, and opening a zipper bag to extract its contents.

\subsection{Facilitating a Hand-Centered Task}
Some tasks that follow self-handovers do not focus on object manipulation, but are instead directed toward the hand itself. For example, one hand may pass a paper towel to the other (and back) in order to wipe water off both hands. In another case involving non-solid objects, one hand applies liquid soap to the other in order to spread the soap across both hands. In these cases, the purpose of the task is not to manipulate the object, but to achieve a desired state of the hands, such as drying or cleaning them. In this type of self-handover, both hands are typically involved in the subsequent action because the focus is on the condition of the hands.

\subsection{Freeing the Hand for a New Task}
\begin{figure}[t]
  \centering
  \includegraphics[width=0.48\textwidth]{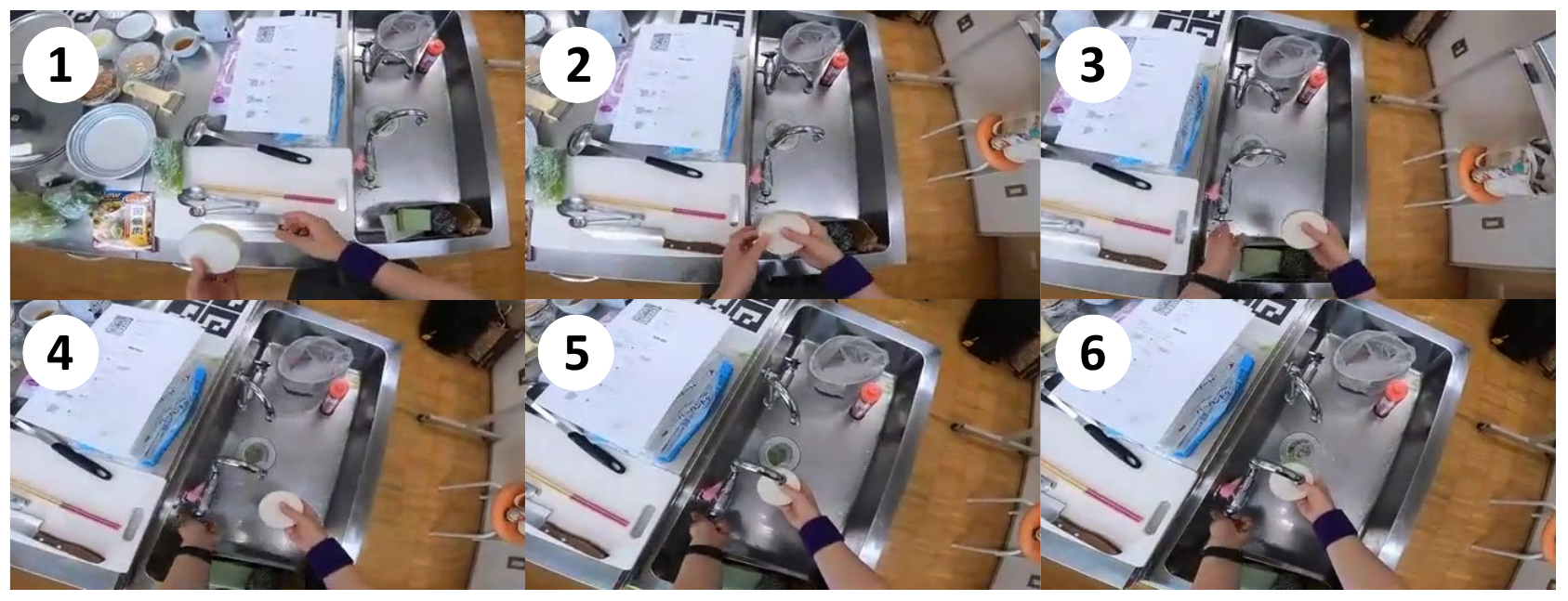}
  \caption{An example that can be interpreted as both ``Freeing the Hand for a New Task'' and ``Simplifying the Subsequent Task.'' After handing over the daikon radish, the left hand turns on the faucet while the right hand holds and rinses the vegetable under running water. The numbers indicate the chronological order of frames.}
  \label{fig:simplifynext}
\end{figure}

In this type of self-handover, the transferring hand proceeds to engage in a task that does not involve the object being passed. Through our observations, we found that many instances of this type can also be interpreted as cases of ``Simplifying the Subsequent Task.'' That is, after the object is handed over, the receiving hand manipulates it while the releasing hand simultaneously performs a different task. For example, after passing a vegetable, the transferring hand may turn on a faucet while the receiving hand rinses it under running water (Fig.~\ref{fig:simplifynext}). This strategy enables parallel processing and appears to be a common behavior in efficient human workflows.

The intention behind such a handover can be interpreted from two perspectives: when focusing on the releasing hand, it serves the purpose of ``Freeing the Hand for a New Task''; when focusing on the receiving hand, it contributes to ``Simplifying the Subsequent Task.'' On the other hand, we also observed cases where the sole purpose was to free the transferring hand. In such cases, the object is intentionally passed to allow the hand to perform a task more effectively---due to factors such as spatial advantage or handedness. Rather than enforcing strict boundaries, our taxonomy is designed to accommodate such fluidity, allowing multiple valid interpretations depending on task context and hand roles.

Unlike other types of self-handover, this category is characterized by a relatively low degree of attention to the transferred object in the immediate subsequent task. Accordingly, in Fig.~\ref{taxonomy}, we highlight this category in blue to distinguish it as a more passive type of self-handover, in contrast to the other categories, which are shown in red, where the object plays an active role in the subsequent task.

\section{Cooking Video Analysis}\label{cooking}
The taxonomy presented in Section~\ref{taxonomy} was constructed based on observations of human cooking actions. Cooking was chosen among various everyday human activities because it involves rich and diverse patterns of object handling and manipulation that can trigger frequent and varied patterns of self-handover. This section introduces the analyzed cooking activities and also presents a preliminary experiment using a VLM as a first step toward automatic classification from videos.

\subsection{Cooking Task}
\begin{figure}[t]
  \centering
  \includegraphics[width=0.40\textwidth]{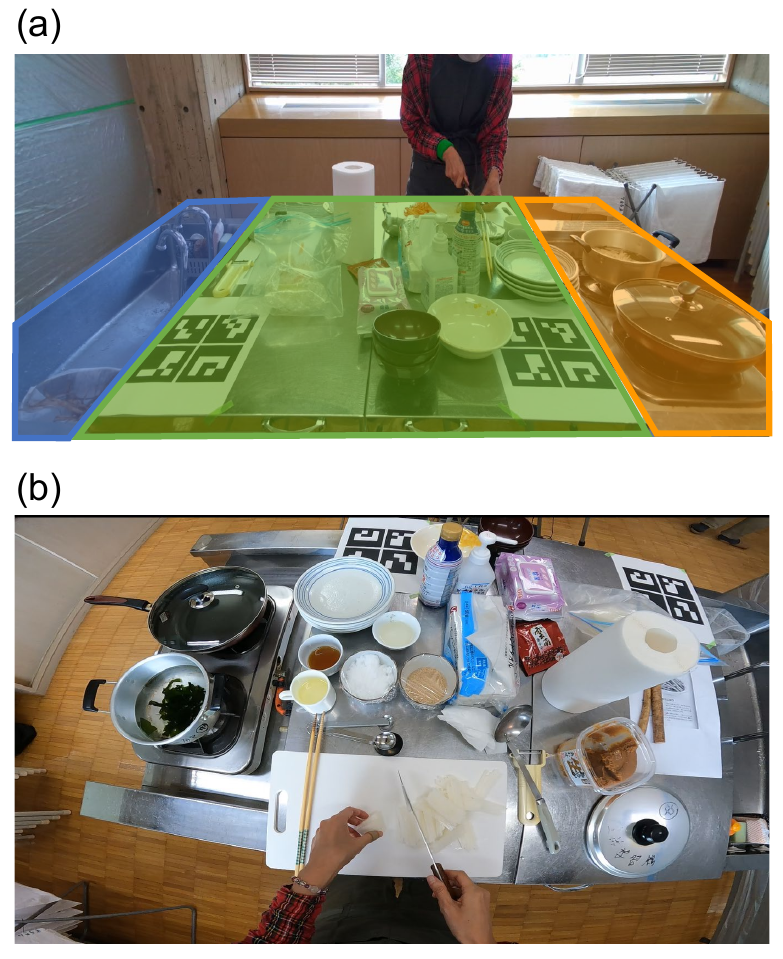}
  \caption{(a) Overview of the kitchen setup, consisting of the washing area (orange), cutting area (green), and heating area (orange). (b) Sample image obtained from the head-mounted GoPro used in the analysis.}
  \label{fig:kitchen}
\end{figure}

The dataset was collected from cooking sessions conducted in a standardized kitchen studio (Fig.~\ref{fig:kitchen}). Twenty-one adult participants, both male and female, each performed two cooking tasks: miso soup and either Kinpira Gobo (stir-fried burdock root, prepared by 11 participants) or Hui Guo Rou (twice-cooked pork, prepared by 10 participants). Participants were non-professional adults with typical home cooking experience. All participants were right-handed. Each participant was provided with detailed recipes and pre-measured ingredients, packaged in clearly labeled transparent plastic bags. The controlled kitchen was composed of three functional zones: a washing area, a cutting area, and a heating area (Fig.~\ref{fig:kitchen}(a)). All necessary utensils, dishes, and ingredients were arranged on the table before the cooking began, and no time limit was imposed.

\subsection{Recording Setup}
Cooking sessions were recorded using a head-mounted GoPro camera (see Fig.~\ref{fig:kitchen}(b) for a captured image). The RGB videos were captured at a resolution of $1920 \times 1080$ with a frame rate of 30 fps, accompanied by audio recordings. Audio data was not utilized in the subsequent analysis. The total duration of the analyzed cooking activities was 46,236 seconds (771 minutes). The average duration per participant was approximately $2202 \pm 528$ seconds (s.d.), with those who prepared Hui Guo Rou averaging $2059 \pm 622$ seconds and those who prepared Kinpira Gobo averaging $2332 \pm 382$ seconds.

\subsection{Data Analysis}
Analyzing long-duration video recordings is labor-intensive. In our previous work~\cite{wake2025open}, we found that VLMs can detect specific actions in long-form videos when prompted with structured visual inputs. Building on this insight, we adopted a semi-automatic pipeline that combines VLM-based video understanding with manual visual inspection to facilitate the analysis:
\begin{enumerate}
\item \textbf{Video Segmentation}: Videos were downsampled to a resolution of $640 \times 360$.

\item \textbf{VLM-Based Action Localization}: Each video was divided into overlapping 1-second segments using a sliding window with a step size of 0.5 seconds. For each segment, 15 evenly sampled frames were extracted, tiled into a single image in temporal order from left to right, and used as input to GPT-4o~\cite{OpenAI} to determine whether the segment contained a self-handover action. At this stage, the goal was not to classify the specific type of self-handover but to identify potential candidates. Accordingly, we used a simple prompt, as illustrated in Fig.~\ref{fig:p_screening}.

\item \textbf{Manual Review and Classification}: All candidate clips were manually reviewed by the authors to remove false positives. During this process, the taxonomy described in Section~\ref{taxonomy} was developed based on the observed patterns, and the remaining clips were annotated accordingly in parallel.
\end{enumerate}

\begin{figure}[t]
  \centering
  \begin{mdframed}[backgroundcolor=gray!10, linecolor=gray!50]
    \scriptsize
    \textcolor{black}{
      The following image sequence shows whether a person is passing something from one hand to the other. Please respond in JSON format using only \{\{'answer': 'yes'\}\} or \{\{'answer': 'no'\}\}.
    }
  \end{mdframed}
  \caption{The prompt for detecting self-handover.}
  \label{fig:p_screening}
\end{figure}

\subsection{Data Summary}
After the semi-automatic data analysis, we extracted 363 video clips that reliably exhibit self-handover actions. Table~\ref{tab:data_profile} summarizes the distribution of self-handover types extracted through the analysis. We grouped the ``Freeing the Hand for a New Task'' and ``Simplifying the Subsequent Task'' types into a single category in this analysis. This decision reflects our observation that ``Freeing the Hand'' rarely occurred in isolation; instead, it almost always co-occurred with ``Simplifying the Subsequent Task'' in the same action sequence. As discussed in Section~\ref{taxonomy}, these two categories often represent two perspectives of the same behavior, and thus were treated as a combined category for annotation consistency. 

The most frequently observed category was ``Performing a Bimanual Task'' (178 clips), followed by ``Freeing the Hand for a New Task / Simplifying the Subsequent Task''  (136 clips). The categories ``Facilitating a Hand-Centered Task'' and ``Adjusting grip'' were less common, with 29 and 20 clips, respectively. This result not only highlights the diversity of self-handover behaviors, but also motivates the use of VLMs for automatic classification.
\begin{table}[ht]
  \centering
  \caption{Distribution of Self-Handover Categories}
  \label{tab:data_profile}
  \begin{tabular}{lc}
    \toprule
    \textbf{Category} & \textbf{Number of Clips} \\ 
    \midrule
    \rowcolor{gray!20}
    Adjusting Grip & 20 \\
    Freeing the Hand for a New Task & \multirow{2}{*}{136} \\ 
    / Simplifying the Subsequent Task & \\
    \rowcolor{gray!20}
    Performing a Bimanual Task & 178 \\ 
    Facilitating a Hand-Centered Task & 29 \\ 
    \bottomrule
  \end{tabular}
\end{table}

\subsection{VLM-based Automatic Classification}
Our proposed taxonomy helps explain why and how humans perform self-handover in a given scene, serving as a basis for learning bimanual robot handover from human demonstration. As a preliminary step toward automatic classification from human demonstrations, we conducted a video classification experiment using GPT-4o. We chose a VLM-based approach for its practical advantages: it requires no task-specific training and offers semantic understanding.

\begin{figure}[t]
  \centering
  \includegraphics[width=0.45\textwidth]{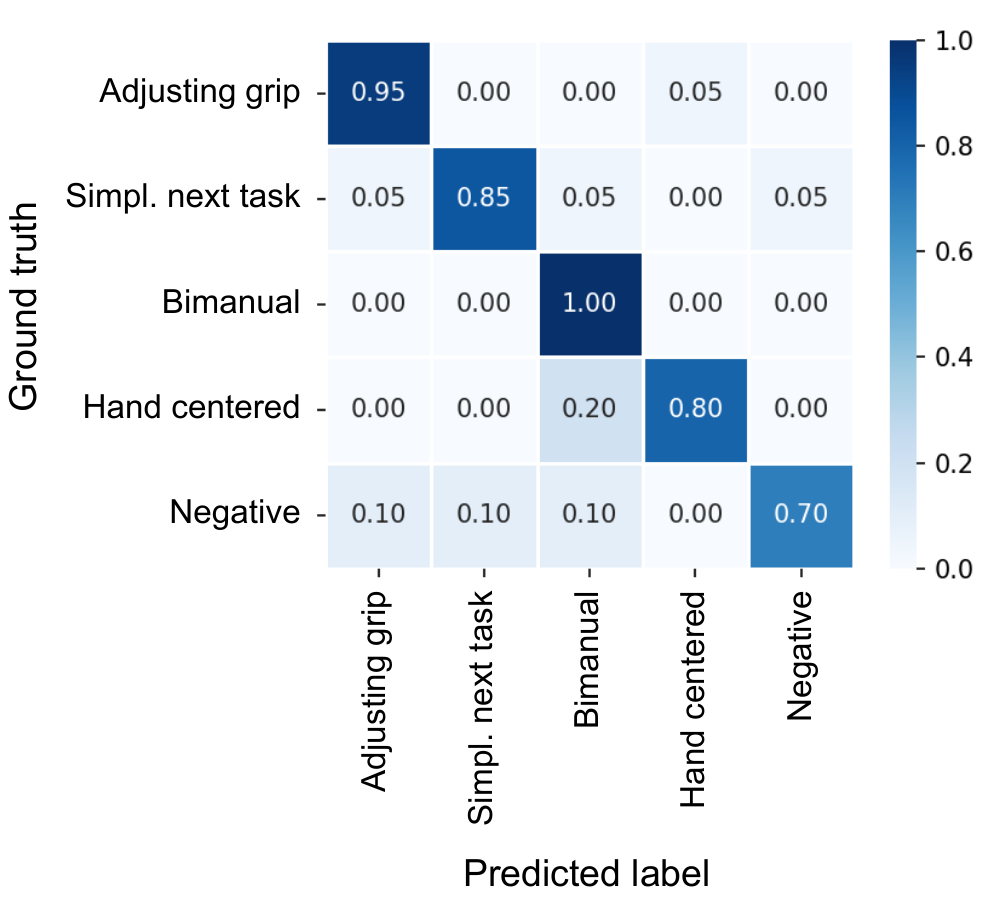}
  \caption{The results of self-handover classification using the VLM are shown as a normalized confusion matrix. Category abbreviations correspond to: Adjusting Grip, Simplifying the Subsequent Task, Performing a Bimanual Task, Facilitating a Hand-Centered Task, and Negative (non-handover actions).}
  \label{fig:matrix}
\end{figure}

We prepared 20 representative clips for each of the four categories in Table~\ref{tab:data_profile}, along with a negative class consisting of bimanual actions without self-handover. For spatial focus of attention, each clip was cropped to a bounding box encompassing both hands, estimated from hand landmarks detected across the entire sequence using a third-party model~\cite{mediapipe}. A structured prompt aligned with the taxonomy (Fig.~\ref{fig:p_role}) was used to classify each video clip. For each clip, 15 evenly sampled frames were extracted and tiled horizontally in temporal order to form a single image, and then used as input to GPT-4o. As shown in the confusion matrix (Fig.~\ref{fig:matrix}), the model achieved reasonable performance with an overall accuracy of 86\%, although the performance for the Negative class was the lowest at 70\%. These results indicate the potential of VLMs for automatic self-handover classification, laying the groundwork for learning structured bimanual behaviors in robotic systems.

\begin{figure}[t]
  \centering
  \begin{mdframed}[backgroundcolor=gray!10, linecolor=gray!50]
    \scriptsize
    \textcolor{black}{
    Your task is to classify the given video clip based on the self-handover taxonomy described below. \\
    \textbf{Self-Handover Taxonomy}\\
    - Type 1: Regrasp …\\
    - Type 2: Simplify\_Next …\\
    - Type 3: Hand\_Centered …\\
    - Type 4: Bimanual …\\
    - Type 5: Non\_handover …\\
    \textbf{Instructions}\\
    - Analyze the given video clip and classify it according to the taxonomy above.\\
    - Choose Type 1-4 only when a self-handover is clearly observed; otherwise, assign Type 5.\\
    - If multiple categories apply, select the one that best reflects the intent.\\
    - Provide a brief justification for your choice.\\
    \textbf{Output Format}\\
    …
    }
  \end{mdframed}
  \caption{The prompt for classifying video clips.}
  \label{fig:p_role}
\end{figure}

\section{Discussion}
\subsection{Use of VLM for automatic classification}
This study presents a preliminary experiment using a VLM to detect and classify self-handover actions from video data. While our main contribution lies in the proposed taxonomy, we included the classification task as a practical demonstration. GPT-4o enabled semantic-level recognition without requiring task-specific data collection or model training. Nonetheless, higher classification accuracy could likely be achieved with dedicated models trained on large-scale annotated datasets.

Accurately recognizing context-dependent self-handover remains a challenging problem. For example, Krebs et al.~\cite{krebs2022bimanual} showed that rule-based methods using hand motion features can be effective for bimanual task classification, but argued that such methods are insufficient when semantic understanding of task goals is required. Self-handover actions also differ from typical action recognition targets---such as sports or household activities---in that they are brief, often completed within a second. 
In this respect, GPT-4o has been shown to recognize short, intent-driven actions from both first-person and third-person viewpoints, even in the absence of explicit temporal context~\cite{wake2025open}.

Based on these considerations, VLMs appears to be well suited for classifying handover behaviors.

\begin{figure}[t]
  \centering
  \includegraphics[width=0.48\textwidth]{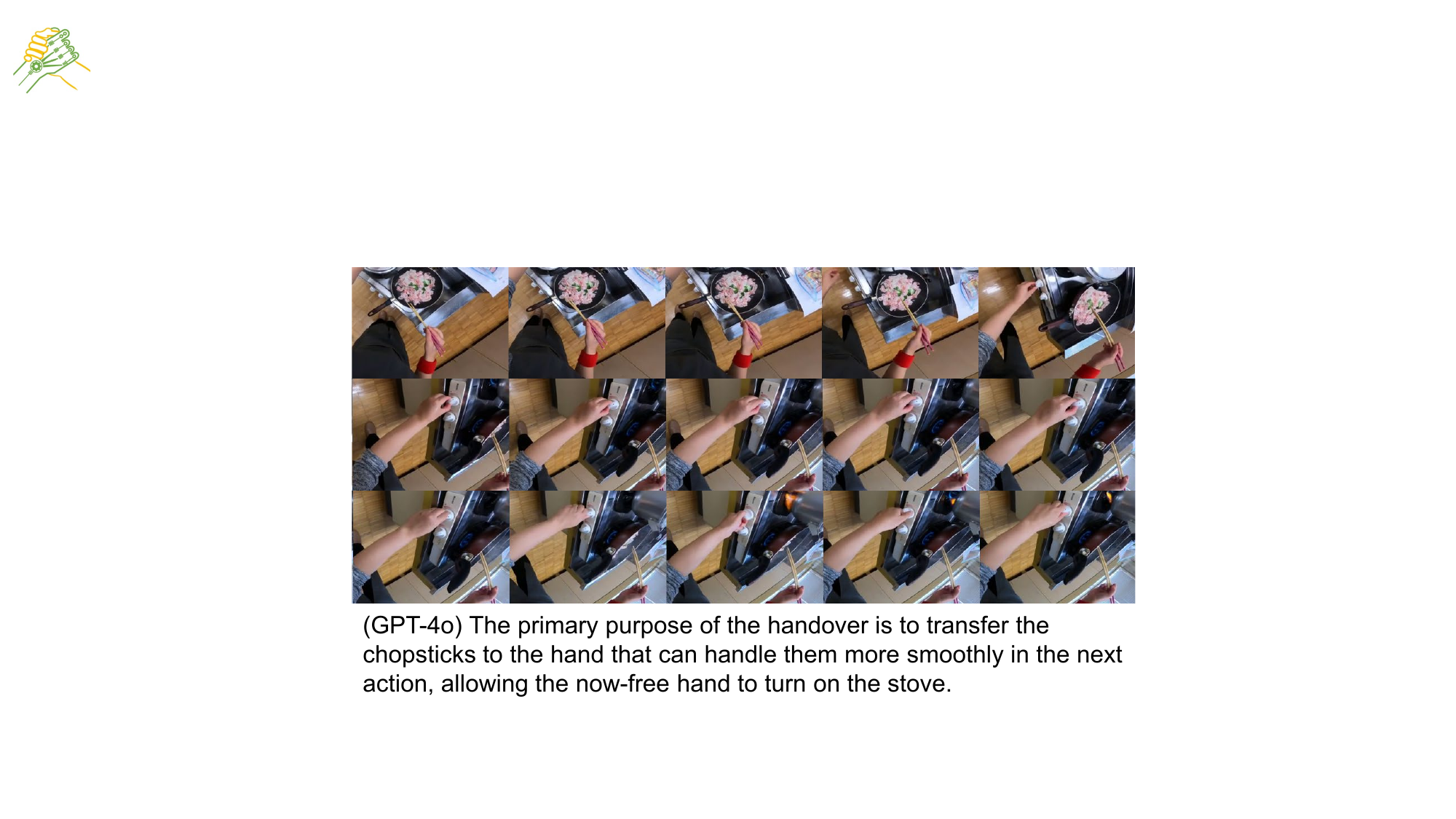}
  \caption{An example where the VLM incorrectly identifies a self-handover in a negative-class video. GPT refers to the visible chopsticks and erroneously describes a handover involving them.}
  \label{fig:hallucination}
\end{figure}

At the same time, VLMs exhibit a notable limitation: they may occasionally produce plausible but incorrect interpretations of a scene---known as hallucinations~\cite{li2023evaluating}. In our experiment, GPT-4o sometimes described nonexistent actions in negative-class clips, even when temporally dense frame sequences clearly showed no handover (Fig.\ref{fig:hallucination}). These errors appeared to stem, in some cases, from object-based inference, highlighting the importance of carefully preprocessing visual data to minimize misleading contextual cues.

\subsection{Self-handover and handedness}
As explained in Section~\ref{taxonomy}, our observations suggest that self-handover is sometimes performed to allow the dominant hand to execute the subsequent task. Prior research by Guiard~\cite{guiard1987asymmetric} and Sainburg~\cite{sainburg2002evidence} has shown that the dominant and non-dominant hands often play asymmetric roles in human bimanual coordination. These findings are further supported by neuroscientific evidence~\cite{johansson2006lateralized}, suggesting that accounting for handedness can facilitate the analysis of self-handover behaviors in humans.

From a robotics perspective, however, handedness may be less critical, unless the object is physically designed for a particular hand. Nevertheless, our taxonomy remains applicable, as it is grounded in the intent behind the subsequent task, regardless of the existence of handedness. This task-oriented perspective allows the taxonomy to inform both human behavior analysis and the development of robotic self-handover strategies.

\section{Conclusion}
We have proposed a taxonomy of self-handover based on the observation that humans strategically perform self-handover actions to facilitate subsequent tasks. Accordingly, our classification framework takes into account both the hand that executes the subsequent task and the object involved. While existing literature has addressed robotic self-handover, no comprehensive framework yet exists for the context-aware selection and execution of varied self-handover types. 
We believe that the proposed taxonomy not only deepens our understanding of human behavior but also offers a practical foundation for robotic systems to perform context-aware and adaptive self-handover. Future work will focus on its integration into robotic learning frameworks to enable autonomous decision-making in diverse, unstructured environments.

\section*{Acknowledgments}
We thank Dr. Midori Otake (Tokyo Gakugei University), Dr. Etsuko Saito (Ochanomizu University), and Dr. Sakiko Yamamoto (Niigata University) for their assistance in collecting the cooking dataset. This study was conceptualized, conducted, and written solely by the authors. We used OpenAI's GPT-4o for language editing and proofreading, with all scientific content and interpretations provided by the authors.

\bibliographystyle{ieeetr}
\bibliography{bib}
\end{document}